\pdfoutput=1
\documentclass[letterpaper, 10 pt, journal, twoside]{IEEEtran}
\usepackage{times}
\usepackage{epsfig}
\usepackage{graphicx}
\usepackage{amsmath}
\usepackage{amssymb}
\usepackage{bbm}
\usepackage{mathtools}
\usepackage{bm}
\usepackage{esvect}

\usepackage{cite}
\usepackage{float}
\usepackage{algorithm}
\usepackage{algorithmicx}
\usepackage{algpseudocode}
\usepackage{mathtools}
\usepackage{url}
\usepackage[table]{xcolor} 
\usepackage[para,online,flushleft]{threeparttable}
\usepackage{etoolbox}
\appto\TPTnoteSettings{\footnotesize}

\usepackage{arydshln}
\usepackage{subcaption}
\usepackage{makecell}
\usepackage{pifont}
\usepackage{multirow}
\usepackage{booktabs}
\usepackage{chngcntr}
\usepackage{float}
\usepackage{diagbox}

\usepackage{amsmath}

\DeclareMathOperator*{\argmin}{arg\,min}

\usepackage[breaklinks=true,bookmarks=false]{hyperref}
\hypersetup{
    colorlinks=true,
    linkcolor=blue,
    filecolor=magenta,
    urlcolor=cyan,
}

\begin{document}

\title{LaneAF: Robust Multi-Lane Detection \\with Affinity Fields}

% Paper headers
\markboth{IEEE Robotics and Automation Letters. Preprint Version. Accepted July, 2021}
{Abualsaud \MakeLowercase{\textit{et al.}}: LaneAF} 
% Use only for final RAL version

% Make room for more info lines in the \author command 
\author{Hala Abualsaud$^{1\dag}$, Sean Liu$^{1\dag}$, David B. Lu$^{1\dag}$, Kenny Situ$^{1\dag}$, Akshay Rangesh$^{1\dag}$, and Mohan M. Trivedi$^{1}$%
\thanks{Manuscript received: March 22, 2021; Revised June 15, 2021; Accepted July 11, 2021.}%Use only for final RAL version
\thanks{This paper was recommended for publication by Editor Cesar Cadena upon evaluation of the Associate Editor and Reviewers' comments.}
\thanks{$^{1}$The authors are with Laboratory for Intelligent and Safe Automobiles, UC San Diego, CA 92092, USA
{\tt\footnotesize \{habualsa, sel118, dblu, ksitu, arangesh, mtrivedi\}@ucsd.edu}}%
\thanks{$^{\dag}$ Equal contribution}
\thanks{Code: \href{https://github.com/sel118/LaneAF}{https://github.com/sel118/LaneAF}}
\thanks{Digital Object Identifier (DOI): see top of this page.}
}
% Use only for final RAL version.

\maketitle
% \pagestyle{empty}
% \thispagestyle{empty}
% Comment or remove these lines for final RAL version.

%%%%%%%%% ABSTRACT
\begin{abstract}
This study presents an approach to lane detection involving the prediction of binary segmentation masks and per-pixel affinity fields. These affinity fields, along with the binary masks, can then be used to cluster lane pixels horizontally and vertically into corresponding lane instances in a post-processing step. This clustering is achieved through a simple row-by-row decoding process with little overhead; such an approach allows LaneAF to detect a variable number of lanes without assuming a fixed or maximum number of lanes. Moreover, this form of clustering is more interpretable in comparison to previous visual clustering approaches, and can be analyzed to identify and correct sources of error. Qualitative and quantitative results obtained on  popular lane detection datasets demonstrate the model's ability to detect and cluster lanes effectively and robustly. Our proposed approach sets a new state-of-the-art on the challenging CULane dataset and the recently introduced Unsupervised LLAMAS dataset.
\end{abstract}

% Keywords appear just beneath the abstract. Use only for final RAL version. 
\begin{IEEEkeywords}
Object Detection, Segmentation and Categorization, Deep Learning for Visual Perception
\end{IEEEkeywords}

%%%%%%%%% BODY TEXT
\section{Introduction}
% Drop letter for first word of the Introduction
\IEEEPARstart{L}{ane} detection is the process of automatically perceiving the shape and position of marked lanes and is a crucial component of autonomous driving systems, directly influencing the guidance and steering of vehicles while also aiding the interaction between numerous agents on the road. As the number of drivers on the roads has increased, autonomous driving systems have received considerable attention in the automotive and tech industries as well as in academia \cite{daily2017self}. According to the Insurance Institute for Highway Safety (IIHS), in the US alone, car accidents claimed 36,560 lives in 2018, underscoring the importance of any technology that can help prevent crashes. 

Since roads commonly have different types of lane lines (solid white, broken white, solid yellow, etc.), each of which have specific implications with regards to how vehicles may interact with them, automated lane detection systems can also help alert drivers when there are changes in lane topology on the road. Furthermore, there are several factors that make lane detection a challenging task. Firstly, there is a wide variety of road infrastructure in use around the world. Additionally, the lane detection system must be able to identify instances where lanes are ending, merging, and splitting.  Finally, the lane detection system must possess the ability to discern worn or unclear lane markings. Precise detection of lanes also enable more robust trajectory prediction of surrounding vehicles; as discussed in \cite{deo2018would}, this is critical for successful path planning in autonomous driving. Therefore, while lane detection is a significant and complex task, it is a key factor in developing any autonomous vehicle system.

%In this paper, lane detection is approached using Deep Layer Aggregation (DLA) as the backbone network because of its accuracy and efficiency; in \cite{dla}, the authors used DLA to achieve best-in-class accuracy on semantic segmentation of the CityScapes dataset. Since CityScapes is a large-scale and challenging dataset with 19 semantic categories, it was believed that DLA could achieve or surpass current state-of-the-art performance in the lane detection task, which involves significantly fewer categories. 

\begin{figure}[t]
\centering
\includegraphics[width=\linewidth]{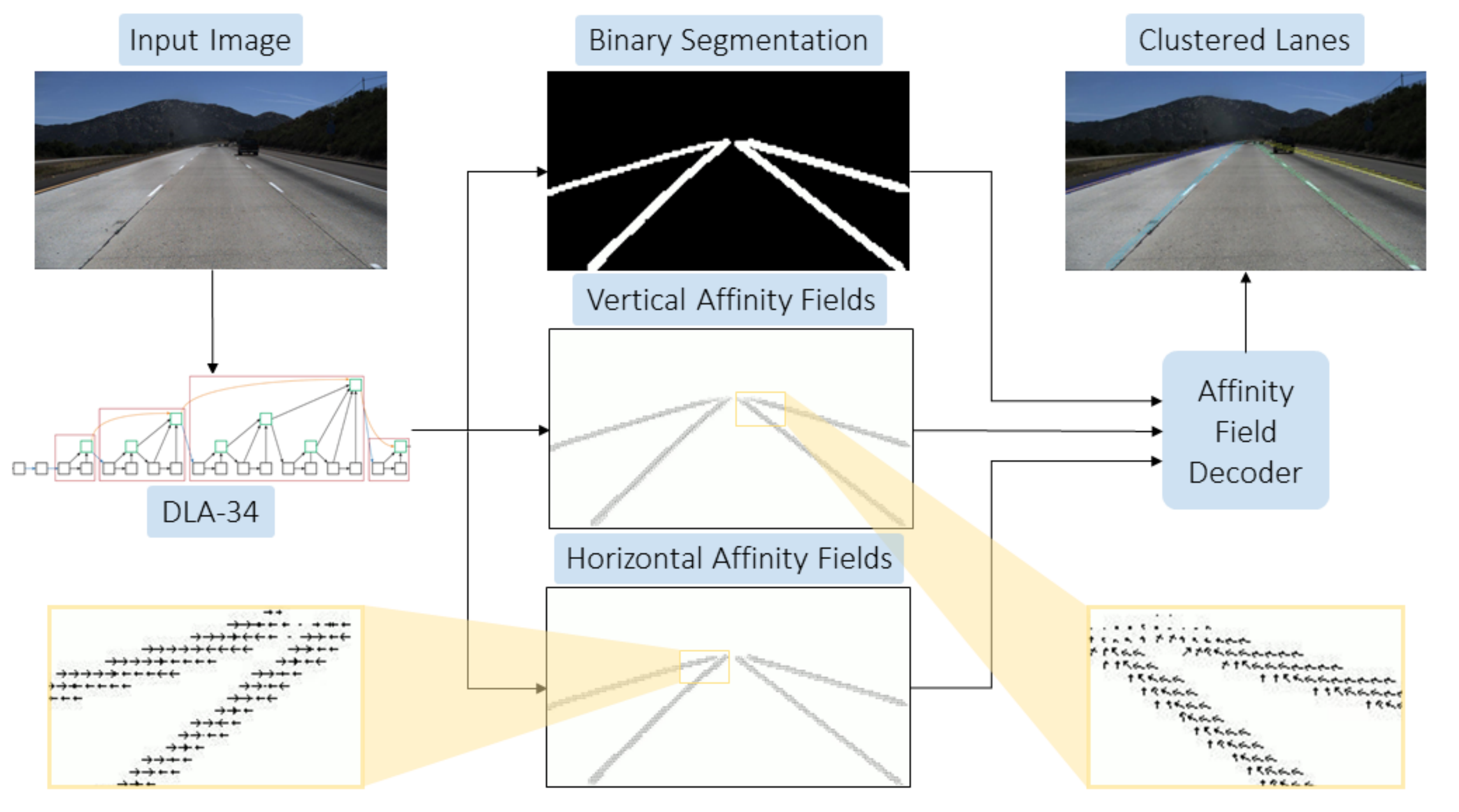}
\caption{In our approach, we propose to train a model that outputs binary segmentation masks and affinity fields, which can then be decoded together to produce multiple lane instances. This is opposed to the standard approach to (anchor-free) lane detection that treats each lane as a separate class and trains a model to perform multi-class segmentation.}
\label{fig:approach}
\end{figure}

While binary classification is used for the detection of lanes in our approach, a limitation of this type of classification is that it produces a single-channel output, which does not allow for the identification of separate lane entities. To dissociate different lane instances, we propose a novel clustering scheme based on affinity fields (see Figure~\ref{fig:approach}). Affinity fields were originally introduced in \cite{affinity} for the purpose of multi-person 2D pose estimation, and are comprised of unit vectors that encode location and orientation. This technique was also used for the detection of hands inside a vehicle, as demonstrated in \cite{yuen2019looking}. In this paper, we have defined two types of affinity fields, the horizontal affinity field (HAF) and vertical affinity field (VAF).
It is these affinity fields that enable unique lane instances to be identified and segmented. Since these affinity fields are present wherever there are foreground lane pixels, they are not bound to a pre-determined number of lanes. The model is therefore agnostic to the number of lanes present on the road. 

The main contributions of this paper are as follows:
\begin{enumerate}
\item We show that using an off-the-shelf convolutional neural network (CNN) backbone~\cite{dla} that intrinsically aggregates and refines multi-scale features can result in superior performance when compared to other bespoke architectures and losses previously proposed for lane detection.
\item We propose affinity fields that are suitable for clustering and associating pixels belonging to amorphous entities like lanes.
\item We detail the procedure and losses to train models that predict binary segmentation masks and affinity fields for the purpose of lane instance segmentation.
\item We introduce efficient methods for generating and decoding such affinity fields into an unknown number of clustered lane instances.
%\item A lane detection method capable of setting a new standard on the CULane Dataset.  
\end{enumerate}

%------------------------------------------------------------------------
\section{Related Research}

Lane detection has traditionally been tackled by feature-based approaches~\cite{satzoda2015enhancing} which then evolved to model-based approaches to detect lane boundaries. However, these are not practical in real world scenarios since they require ideal road scenes to work effectively. Currently, data-driven approaches are commonly used to detect both lane boundaries as well as lane regions. While several shortcomings of the traditional lane detection methods (i.e. lane segmentation via hand-crafted features) have been resolved with more robust methods in recent years, there is still room for improvement. In more recent times, deep learning and large-scale datasets have provided solutions to many of these issues. However, lane detection in unconstrained environments and complex scenarios remain a challenge.
% Furthermore, deep learning helped to introduce a new era of lane detection in recent years; it proved an obvious improvement in the robustness of the lane detection problem. A lot of new approaches use deep learning to tackle lane detection.

Lane detection nowadays is typically modelled as a semantic segmentation problem to extract features using deep learning methods. New approaches tackle lane detection as a multi-class segmentation problem, where each lane forms a separate class. Some of these approaches include: \cite{enet}, \cite{sensors}, \cite{robust}, \cite{gan}, \cite{pan2018spatial}, and \cite{fastDraw}. 
% It achieved an accuracy of 96.53$\%$ with the TuSimple dataset. 
In \cite{robust}, the authors combine a recurrent neural network (RNN) with a CNN for lane prediction and detection. The use of an embedding loss was introduced in \cite{gan} which uses generative adversarial networks (GANs) to better preserve the structure of lanes and to mitigate the problem of complex post-processing for the output of semantic segmentation; 96$\%$ accuracy on the TuSimple dataset was obtained. In \cite{fastDraw}, a sequential prediction network has been used to avoid heuristic-based clustering post-processing. Another network architecture was presented in \cite{leastSquare} with two elements: a deep network which generates weighted pixel coordinates in addition to a differentiable weighted least-squares fitting module. In \cite{sad}, the authors introduced Self Attention Distillation (SAD) loss to avoid models that propagate data sequentially and to decrease inference time. However, the fully connected layer that the SAD model employs is computationally expensive and cannot adapt to any number of lanes. 

Other lane detection approaches choose to first perform binary segmentation of all lanes, followed by a clustering stage to separate each individual lane instance as in \cite{post}, \cite{cascnn}, and \cite{ko2021key}. Instance segmentation is usually approached with the use of complex pipelines; however, many powerful approaches and research were put to come up with better performance techniques including the approach presented in \cite{bai2017deep}, where they used an end-to-end convolutional neural network to tackle the problem that was inspired by the classical watershed transform. Another method toward instance segmentation was based on using a fully convolutional network to predict semantic labels along with depth and an instance-based encoding. This was implemented by using each pixel’s direction toward its corresponding instance center; with the help of low-level computer vision techniques, impressive scene understanding by predicting pixel-wise depth, semantics, and instance-level direction cues was achieved \cite{uhrig2016pixel}. Lane detection is posed as an instance segmentation problem in \cite{post} so that each lane can be detected in an end-to-end manner, adapting to changing numbers of lanes on the road. In \cite{cascnn}, a combination of instance segmentation and classification was used as an end-to-end deep learning real-time method to avoid reliance on two-step detection networks. Although recent methods of lane detection show high accuracy when applied to the popular published datasets, some drawbacks of these current methods are that they are not robust when encountering occlusion and that they require a fixed number of lanes in a scene; thus, they cannot work for a random number of lanes present on the road. Acknowledging this problem in \cite{ko2021key}, the authors use a key points estimation approach to allow for lane detection of an arbitrary numbers of lanes regardless of orientation. 

More recently, some approaches have modelled lane detection as an anchor-based object detection problem such as \cite{mulFeature}, \cite{end}, \cite{tabelini2020keep}, \cite{spatiTemporal}, and \cite{satzoda2015enhancing}.
% The authors in \cite{pan2018spatial} introduced a spatial CNN with learned spatial kernels. This method has improved performance over conventional CNN methods since it provides spatial information by computing slice by slice convolution in feature maps, enabling information to be transferred between pixels within each layer. 
In \cite{spatiTemporal}, a spatio-temporal deep learning method was proposed to mitigate the errors that can occur when experiencing harsh weather or other complex problems in the road, jeopardizing the accuracy of detecting a lane in the scene.  Meanwhile, in \cite{mulFeature}, lane markers were tracked temporally. Additionally, \cite{tabelini2020keep} presents an anchor-based single-stage deep lane detection model using anchors for feature pooling. In \cite{end}, the authors developed 3D-LaneNet, a network that predicts the 3D layout of lanes using a single image. A combination of LiDAR and camera sensors were used in \cite{snesor} for their network to obtain accurate lane detection in 3D space directly. 

% The methods in this paper were inspired by \cite{affinity}, which presented an approach to 2D pose estimation of multiple people in an image through Part Affinity Fields (PAFs). The technique introduced takes an input image, passes it to a two-branch CNN, obtains the confidence maps to detect body parts, utilizes PAFs for parts association, parses using a greedy algorithm, and finally assembles them into the final image with estimated poses. The main takeaway from \cite{affinity} is that parsing based on PAFs adds robustness with regards to part detection and association. 

% first algorithm
\begin{algorithm}[t]
\small
\caption{Creating affinity fields from ground truth data}\label{alg:generating}
\hspace*{\algorithmicindent} \textbf{Inputs:} \\
\hspace*{\algorithmicindent} \hspace*{\algorithmicindent}$SEG (H\times W)$: ground truth segmentation\\
\hspace*{\algorithmicindent} \hspace*{\algorithmicindent}$l_{max}$: maximum number of lanes\\
\begin{algorithmic}[H]

\State $HAF, VAF \gets zeros(H, W, 2)$ \Comment{initialize affinity fields}
\For{$l \gets 1$ \textbf{to} $L$} \Comment{go through each lane}
\State $prev\_cols \gets nonzero(SEG[H, :] == l)$ \Comment{initialize}
\State /* \textit{row-by-row, from bottom to top} */
\For{$y \gets H-1$ \textbf{to} $1$}
\State $cols \gets find(SEG[row, :] == l)$ \Comment{find lane pixels}
\State /* \textit{horizontal affinity field} */
\For{$x$ \textbf{in} $cols$}
% \If{$col < mean(cols)$}
% \State $HAF[row, col, 0] \gets 1.0$ \Comment{points to the right}
% \ElsIf{$col > mean(cols)$}
% \State $HAF[row, col, 0] \gets -1.0$ \Comment{points to the left}
% \Else
% \State $HAF[row, col, 0] \gets 0.0$
% \EndIf
\State $HAF[y, x] \gets \vv{H}_{gt}(x, y)$ \Comment{Eq.~\ref{eq:haf_create}}
\EndFor
\State /* \textit{vertical affinity field} */
\For{$x$ \textbf{in} $prev\_cols$}
% \State $\mathbf{v} \gets [mean(cols) - prev\_col, -1]$ 
% \State \Comment{points to the mean in the row above}
% \State $\mathbf{v}_{norm} \gets \mathbf{v}/||\mathbf{v}||$ \Comment{unit vector}
% \State $VAF[row+1, prev\_col, 0] \gets \mathbf{v}[0]$
% \State $VAF[row+1, prev\_col, 1] \gets \mathbf{v}[1]$
\State $VAF[y+1, x] \gets \vv{V}_{gt}(x, y+1)$ \Comment{Eq.~\ref{eq:vaf_create}}
\EndFor
\State $prev\_cols \gets cols$
\EndFor
\EndFor
\State \textbf{return} $HAF, VAF$
\end{algorithmic}
\end{algorithm}

% second algorithm
\begin{algorithm}[ht]
\small
\caption{Decoding predicted affinity fields into lanes}\label{alg:decoding}
\hspace*{\algorithmicindent} \textbf{Inputs:} \\
\hspace*{\algorithmicindent} \hspace*{\algorithmicindent}$BW (H\times W)$: binary segmentation mask\\
\hspace*{\algorithmicindent} \hspace*{\algorithmicindent}$HAF (H\times W\times 2)$: horizontal affinity field\\
\hspace*{\algorithmicindent} \hspace*{\algorithmicindent}$VAF (H\times W\times 2)$: vertical affinity field\\
\hspace*{\algorithmicindent} \hspace*{\algorithmicindent}$\tau$: clustering threshold\\
\begin{algorithmic}[h]

\State $SEG \gets zeros(H, W)$ \Comment{initialize segmentation output}
\State $lane\_end\_points \gets []$ 
\State \hspace*{\algorithmicindent} \Comment{keeps track of the latest points added to each lane}
\State $L \gets 0$ \Comment{initialize number of lanes to 0}
\State /* \textit{row-by-row, from bottom to top} */
\For{$y \gets H$ \textbf{to} $1$}
\State $cols \gets find(BW[row, :] > 0)$ \Comment{find foreground pixels}
\State /* \textit{cluster horizontally} */
\State $clusters \gets []$
% \State $prev\_col \gets cols[0]$
% \State $cluster \gets []$
\For{$x$ \textbf{in} $cols$}
% \State $cur\_dir \gets HAF[row, col, 0]$
% \State $prev\_dir \gets HAF[row, prev\_col, 0]$
% \If{$cur\_dir>0$ \textbf{and} $prev\_dir<0$} \Comment{end cluster}
% \State $clusters.append(cluster)$
% \State $cluster \gets []$
% \State $cluster.append(col)$
% \Else \Comment{continue cluster}
% \State $cluster.append(col)$
% \EndIf
% \State $prev\_col \gets col$
\State $clusters.update(c^*_{haf}(x, y))$ \Comment{Eq.~\ref{eq:haf_decode}}
\EndFor

\State /* \textit{assign clusters to existing lanes} */
%second nested loop
\For{$l \gets 1$ \textbf{to} $L$} 
% \State $pred\_points \gets add\_vafs(points, VAF)$
% \State \hspace*{\algorithmicindent} \hspace*{\algorithmicindent} \Comment{add VAF vectors to points}
% \State $cluster, e \gets find\_closest(pred\_points, clusters)$
% \State \hspace*{\algorithmicindent} \hspace*{\algorithmicindent} \Comment{find cluster closest to predicted points, and \hspace*{\algorithmicindent} \hspace*{\algorithmicindent} \hspace*{\algorithmicindent} \hspace*{\algorithmicindent} corresponding error}
\If{$d^*(l) <= \tau$} \Comment{error less than threshold (Eq.~\ref{eq:vaf_decode_2})}
\State $lane\_end\_points[l] \gets c^*_{vaf}(l)$ \Comment{Eq.~\ref{eq:vaf_decode_1},Eq.~\ref{eq:vaf_decode_3}} 
\State \hspace*{\algorithmicindent} \hspace*{\algorithmicindent} \Comment{update latest points added to lane}
\For{$x$ \textbf{in} $c^*_{vaf}$(l)}
\State $SEG[y, x] \gets l$ \Comment{assign cluster to lane $l$}
\EndFor
\EndIf
\EndFor

\State /* \textit{spawn new lanes with unassigned clusters} */
\For{$cluster$ \textbf{in} $clusters$}
\If{$cluster$ \textbf{is not assigned}}
\State $L \gets L + 1$
\State $lane\_end\_points[L] \gets cluster$ 
\EndIf
\EndFor

\EndFor
\State \textbf{return} $SEG$
\end{algorithmic}
\end{algorithm}

%-------------------------------------------------------------------------
\section{Methodology}

% \begin{figure*}[ht]
% \centering
% \includegraphics[width=0.8\linewidth]{model_arch.pdf}
% \caption{LaneAF architecture: after an input image has passed through DLA-34, three separate outputs are generated: a binary segmented image, vertical affinity fields (VAFs), and horizontal affinity fields (HAFs), all of which are combined in our decoder to produce instance segmentation results.}
% \label{fig:model_arch}
% \end{figure*}

Our proposed methodology involves a feed-forward CNN that is trained to predict binary lane segmentation masks and per-pixel affinity fields. More specifically, the model is trained to predict two affinity fields, which we call the horizontal affinity field (HAF) and vertical affinity field (VAF), respectively. Affinity fields can be thought of as vector fields that map any 2D location on the image plane to a unit vector in 2D.
A unit vector in the VAF encodes the direction in which the next set of lane pixels above it is located. On the other hand, a unit vector in the HAF points toward the center of the lane in the current row, thereby allowing us to cluster lanes of arbitrary widths. These two affinity fields, in conjunction with the predicted binary segmentation, can then be used to cluster foreground pixels into lanes as a post-processing step. In the next few subsections, we discuss each individual block in our proposed approach.

\subsection{Network Backbone}

Recent lane detection approaches have made use of a variety of backbone architectures, but most popular among them are usually the ResNet family of architectures~\cite{resnet}, ENet~\cite{enet}, and ERFNet~\cite{romera2017erfnet}. Although these architectures have proven benefits across a variety of tasks, we believe that more recent developments in the field can be leveraged for lane detection. To this end, we make use of the DLA-34 backbone presented in~\cite{dla}.

The DLA family of models make use of deep layer aggregation, which unifies semantic and spatial fusion for better localization and semantic interpretation. In particular, this architecture extends densely connected networks~\cite{huang2017densely} and feature pyramid networks with hierarchical and iterative skip connections that deepen the representation and refine resolution. They employ two forms of aggregation: iterative deep aggregation (IDA), focusing on fusing resolutions and scales, and hierarchical deep aggregation (HDA), focusing on merging features from all modules and channels. These architectures also incorporate deformable convolution operations~\cite{dai2017deformable} that can adapt the spatial sampling grid for convolutions based on their inputs. We believe these are desirable properties for the tasks of lane detection and instance segmentation.

\subsection{Affinity Fields}

% We implemented a parallel branch to the binary segmentation branch that predicts the HAF and VAF for instance segmentation of different lane markings. Specifically, affinity fields are unit vectors that point towards the object of interest. In lane detection, for the VAF, this consists of each lane pixel pointing toward the mean location of the next row's lane pixels. For the HAF, this consists of the lane pixels comprising the width of the identified lane line pointing toward the center of that lane line. Our approach can be summarized in the following two steps: 
In addition to binary lane segmentation masks, our model is trained to predict horizontal and vertical affinity fields (HAFs and VAFs respectively). For any given image, the HAF and VAF can be thought of as vector fields $\vv{H}(\cdot, \cdot)$ and $\vv{V}(\cdot, \cdot)$, that assign a unit vector to each $(x, y)$ location in the image. As we alluded to earlier, the HAF enables us to cluster lane pixels horizontally and the VAF vertically. With the predicted affinity fields and binary mask, clustering lane pixels is achieved through a simple row-by-row decoding process from bottom to top. The rest of this subsection provides details on how to create such affinity fields using the ground truth and how to use the predicted affinity fields to decode individual lanes.

\begin{figure*}[t]
\captionsetup[subfigure]{justification=centering}
  	\centering
  	\begin{subfigure}[t]{0.45\linewidth}
		\centering
		\includegraphics[width=\linewidth]{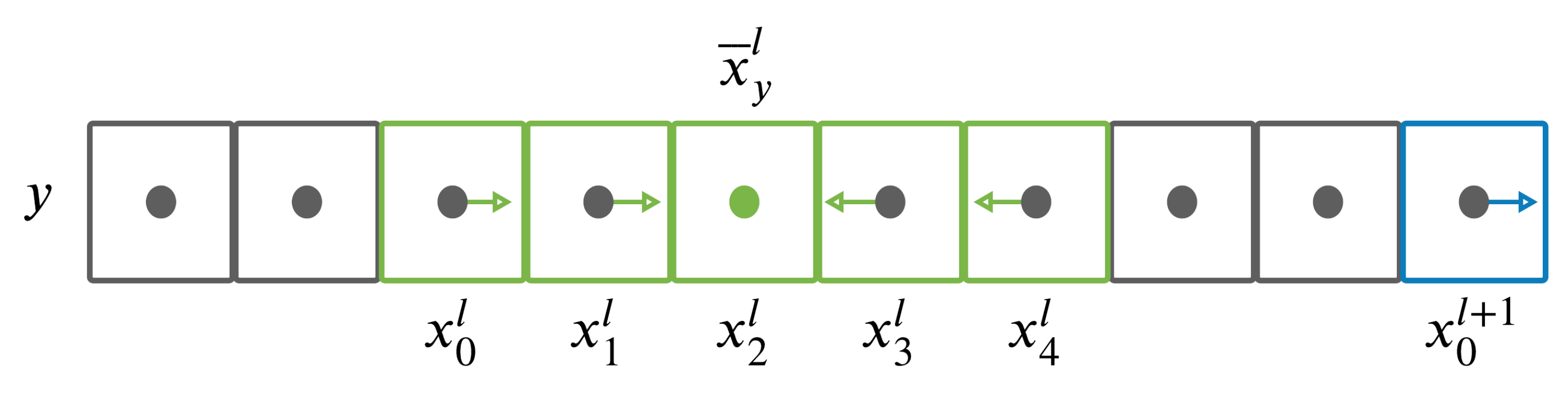}
		\caption{HAF creation during training}
		\label{fig:haf_train}
	\end{subfigure}\hspace{4mm}%
~  	
	\begin{subfigure}[t]{0.45\linewidth}
		\centering
		\includegraphics[width=\linewidth]{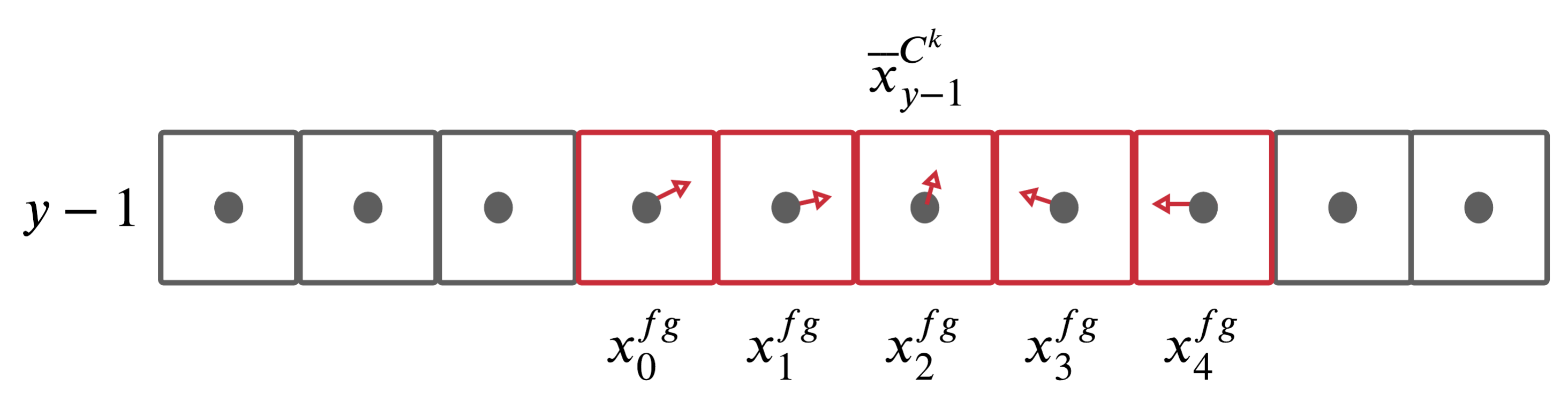}
		\caption{HAF decoding during testing}
		\label{fig:haf_test}
	\end{subfigure}
	
	\begin{subfigure}[t]{0.45\linewidth}
		\centering
		\includegraphics[width=\linewidth]{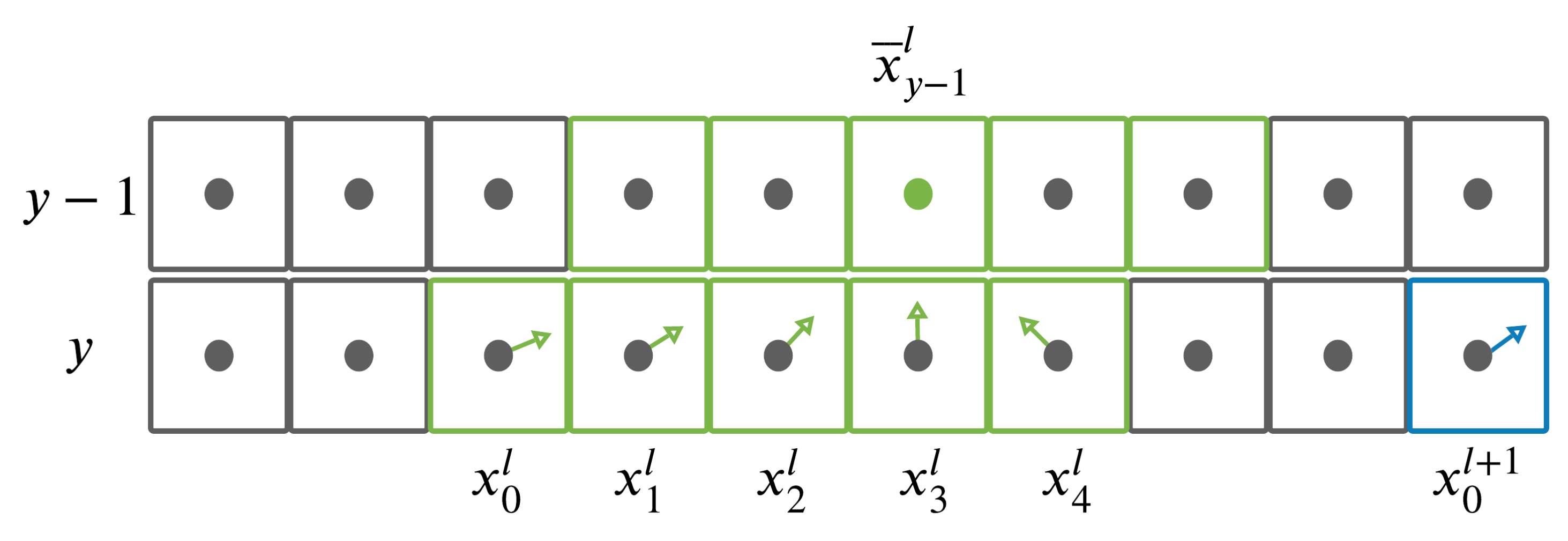}
		\caption{VAF creation during training}
		\label{fig:vaf_train}
	\end{subfigure}\hspace{4mm}%
~  	
	\begin{subfigure}[t]{0.45\linewidth}
		\centering
		\includegraphics[width=\linewidth]{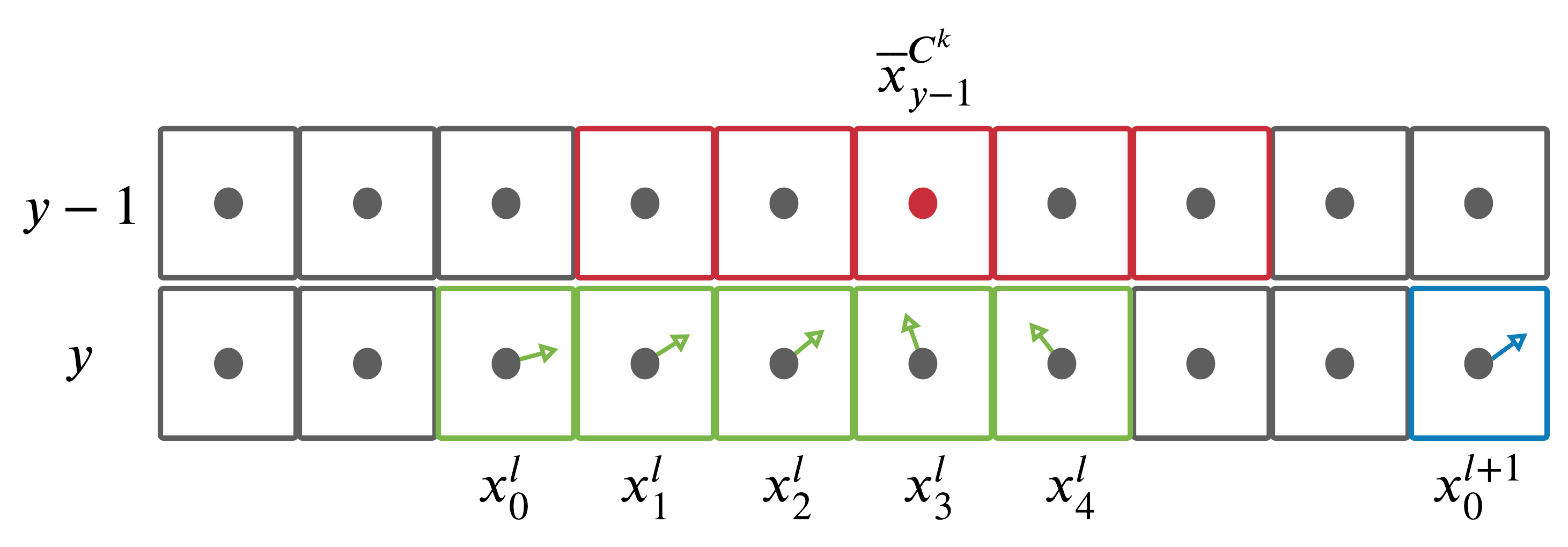}
		\caption{VAF decoding during testing}
		\label{fig:vaf_test}
	\end{subfigure}
	\caption{Illustrations of HAF and VAF creation and decoding processes during training and testing respectively.}
	\label{fig:haf_vaf}
\end{figure*}

\noindent\textbf{Creating HAFs and VAFs:} Affinity fields are created using ground truth segmentation masks on the fly as detailed in Algorithm~\ref{alg:generating}. This proceeds row-by-row, from bottom to top.

For any row $y$ in the image, the HAF vectors are computed for each lane point $(x_i^l, y)$ using the ground truth vector field mapping $\vv{H}_{gt}(\cdot, \cdot)$ as follows:
\begin{equation}~\label{eq:haf_create}
\begin{split}
\vv{H}_{gt}(x_i^l, y) 
&= \bigg(\frac{\overline{x}^l_{y} - x_i^l}{|\overline{x}^l_{y} - x_i^l|}, \frac{y - y}{|y - y|}\bigg)^\intercal \\
&= \bigg(\frac{\overline{x}^l_{y} - x_i^l}{|\overline{x}^l_{y} - x_i^l|}, 0\bigg)^\intercal,
\end{split}
\end{equation}
where $\overline{x}^l_{y}$ is the mean $x$-coordinate of all points belonging to lane $l$ in row $y$. This process is illustrated in Figure~\ref{fig:haf_train}, where pixels in green and blue represent points belonging to lanes $l$ and $l+1$ respectively.

Similarly, the VAF vectors are computed for each lane point $(x_i^l, y)$ in row $y$ using the ground truth vector field mapping $\vv{V}_{gt}(\cdot, \cdot)$ as follows:
\begin{equation}~\label{eq:vaf_create}
\begin{split}
\vv{V}_{gt}(x_i^l, y) 
&= \bigg(\frac{\overline{x}^l_{y-1} - x_i^l}{|\overline{x}^l_{y-1} - x_i^l|}, \frac{y-1 - y}{|y-1 - y|}\bigg)^\intercal \\
&= \bigg(\frac{\overline{x}^l_{y-1} - x_i^l}{|\overline{x}^l_{y-1} - x_i^l|}, -1\bigg)^\intercal,
\end{split}
\end{equation}
where $\overline{x}^l_{y-1}$ is the mean $x$-coordinate of all points belonging to lane $l$ in row $y-1$. This process is illustrated in Figure~\ref{fig:vaf_train}, where pixels in green represent points belonging to lanes $l$. Note that unlike the HAF, unit vectors in the VAF point to the mean location of the lane in the previous row.

\noindent\textbf{Decoding HAFs and VAFs:} After a model is trained to predict the HAFs and VAFs detailed above, a decoding procedure is carried out to cluster foreground pixels into lanes during testing. This procedure is presented in Algorithm~\ref{alg:decoding}, and similarly operates row-by-row, from bottom to top.

Assuming $\vv{H}_{pred}(\cdot, \cdot)$ is the vector field corresponding to the predicted HAF, foreground pixels in a row $y-1$ are first assigned to clusters based on the following rule:
\begin{equation}~\label{eq:haf_decode}
c^*_{haf}(x_i^{fg}, y-1) = 
\begin{cases} 
C^{k+1} &\mbox{if } \begin{aligned}[t]\vv{H}_{pred}(x_{i-1}^{fg}, y-1)_0 \leq 0\\\wedge\ \vv{H}_{pred}(x_{i}^{fg}, y-1)_0 > 0,\end{aligned} \\
C^{k} & \mbox{otherwise,}
\end{cases}
\end{equation}
where $c^*_{haf}(x_i^{fg}, y-1)$ denotes the optimal cluster assignment for a foreground pixel $(x_i^{fg}, y-1)$; $C^k$ and $C^{k+1}$ denote two different clusters indexed by $k$ and $k+1$ respectively. This assignment is illustrated in Figure~\ref{fig:haf_test}, where pixels in red are assigned the same cluster.

Next, these horizontal clusters are assigned to existing lanes indexed by $l$ using the vector field $\vv{V}_{pred}(\cdot, \cdot)$ corresponding to the VAF as follows:
\begin{equation}~\label{eq:vaf_decode_1}
c^*_{vaf}(l) = \underset{C^k}{\argmin}\ d^{C^k}(l),
\end{equation}
where
\begin{equation}~\label{eq:vaf_decode_2}
d^*(l) = \underset{C^k}{\min}\ d^{C^k}(l).
\end{equation}
% \begin{equation}~\label{eq:vaf_decode_1}
% c^*_{vaf}(l) = 
% \begin{cases} 
% \underset{C^k}{\argmin}\ d^{C^k}(l) &\mbox{if } d^*(l) = \underset{C^k}{\min}\ d^{C^k}(l) \leq \tau \\
% C^{k_{tot}+1} & \mbox{otherwise,}
% \end{cases}
% \end{equation}
Here, $d^{C^k}(l)$ denotes the error of associating cluster $C^k$ to an existing lane $l$:
\begin{equation}~\label{eq:vaf_decode_3}
\begin{split}
d^{C^k}(l) = &\frac{1}{N^l_y} \sum_{i=0}^{N^l_y-1} \bigg|\bigg|(\overline{x}^{C^k}, y-1)^\intercal -(x_i^l, y)^\intercal \\
& -\vv{V}_{pred}(x_i^l, y)\cdot||(\overline{x}^{C^k}, y-1)^\intercal - (x_i^l, y)^\intercal||\bigg|\bigg|,
\end{split}
\end{equation}
where $N^l_y$ are the number of pixels belonging to lane $l$ in row $y$.
We illustrate this process in Figure~\ref{fig:vaf_test}, where the cluster in red is assigned to the existing lane in green. By repeating the above steps row-by-row starting from the bottom and working to the top, we are able to assign every foreground pixel to their respective lanes.

% With these two steps, we use an $L2$ norm (same as in \cite{affinity}) to compute the losses between our predicted affinity fields and the ground truth affinity fields that were generated based on the ground truth label of the images. For LaneAF, we combine the DLA backbone with our affinity fields to predict the segmented and uniquely identified lanes. Our entire model architecture is displayed in Figure \ref{fig:approach}. 

\subsection{Losses}
To train the proposed model, we use a separate loss at each prediction head. For our binary segmentation branch, we used weighted binary cross-entropy loss, a standard loss for imbalanced binary segmentation tasks. The raw logits produced by the model are first passed through a sigmoid activation for normalization. The loss is then calculated as: 
\begin{equation}~\label{eq:bce}
    L_{BCE} = -\frac{1}{N}\sum_i \bigg[w \cdot t_i \cdot log(o_i)+(1-t_i) \cdot log(1-o_i)\bigg],
\end{equation}
where $t_i$ is the target value for the pixel $i$ and $o_i$ is the sigmoid output. Since this is an unbalanced segmentation task, a weight $w$ was used to increase penalization for foreground pixels. To further account for the imbalanced dataset, an additional intersection over union loss was used for the segmentation branch:
\begin{equation}
L_{IoU} = \frac{1}{N} \sum_i \bigg[1-\frac{t_{i} \cdot o_{i}}{{t_{i}+o_{i}}-t_{i}\cdot o_{i}}\bigg].
\end{equation}

For the affinity field branches of the model, a simple $L1$ regression loss was applied only to the foreground locations of both the vertical and horizontal affinity fields:
\begin{equation}
L_{AF} = \frac{1}{N_{fg}} \sum_i \bigg[|t^{haf}_{i}-o^{haf}_{i}| + |t^{vaf}_{i}-o^{vaf}_{i}|\bigg].
\end{equation}

The total loss applied to the model is a simple summation of the individual losses:
\begin{equation}
L_{total}=L_{BCE} + L_{IoU} + L_{AF}.
\end{equation}

%-------------------------------------------------------------------------
\section{Experimental Evaluation}

\subsection{Implementation Details}
Our backbone architecture (DLA-34) is a fully convolutional network that does not retain the original resolution, but rather downsizes the outputs by a factor of 4; thus, we re-scaled the input images to one-half their original resolution during run-time and reshaped the ground truth affinity fields and segmentation masks to one-eighth the original resolution (accounting for the model's downsizing factor). This has the added benefit of making our decoding process faster since we now process only an eighth of the original rows. The decoding time typically depends on the number of lanes, the quality of the outputs produced by the model, and the output size. On average, it takes about 15-20ms on a modern CPU without any code optimizations. However, since this an entirely CPU-based operation, it should not affect the overall latency of the approach. We also make use of random rotations, crops, scales and horizontal flips during training.

We use the Adam optimizer as our solver with a learning rate of $0.0001$, weight decay of $0.001$, and train for a total of 40 epochs. We also employ a scheduler that reduces the learning rate by a factor of 5 every 10 epochs.
The weight $w$ for the loss in Eq.~\ref{eq:bce} was set to 9.6 because there are approximately 9.6 times as many background pixels than there are foreground (lane) pixels in most public datasets.
To avoid overfitting, early stopping was implemented by retaining the model parameters that best performed on the validation set. Using a single GTX Titan X Maxwell GPU, training our model on the CULane dataset until convergence (about 25-30 epochs) takes 2-3 days. Significant speedup can be obtained by using more modern GPUs and by employing multiple GPUs when available. 

\subsection{Datasets}
To train and benchmark our proposed approach, we make use of the popular TuSimple, CULane~\cite{pan2018spatial}, and LLAMAS~\cite{llamas} datasets. TuSimple %contains 3,626 annotated training video clips and 2,782 clips for testing. It 
features good and fair weather conditions in various daytime lighting and traffic conditions, employing highways with up to five lanes. Meanwhile, CULane contains significantly more data % with 88,880 annotated training video clips and 34,680 clips for testing. 
and also divides test images into nine categories that contain more complex scenarios, including images with challenging lighting conditions. Finally, the LLAMAS dataset is a newer dataset with a sizeable amount of images all obtained using highway recordings and generated from an automated labeling pipeline. A summary of all datasets is compiled in Table \ref{dataset_table}. %contains 58,269 labeled lane marker clips for training and 20,929 clips for testing. All images were 

\begin{table}[t]
\centering
\caption{Attributes of popular lane detection datasets}
\begin{tabular}{cccc}
    \hline
	\ Dataset & TuSimple & CULane & LLAMAS\\
	\hline\hline
	\multicolumn{1}{r|}{\# Frames} & 6,408 & 133,325 & 100,042 \\
	\hline
	\multicolumn{1}{r|}{Train} & 3,268 & 88,880 & 58,269 \\ 
	\hline
	\multicolumn{1}{r|}{Validation} & 358 & 9,675 & 20,844\\
	\hline
	\multicolumn{1}{r|}{Test} & 2,782 & 34,680 & 20,929\\
	\hline
	\multicolumn{1}{r|}{Resolution} & $1280\times720$ & $1640\times590$ & $1280\times717$\\
	\hline
	\multicolumn{1}{r|}{Road Type} & highway & urban, rural, highway & highway\\
	\hline
	% \multicolumn{1}{r|}{\# Lanes} & $\leq$ 4 & $\ngtr$ 5\\
	%\hline
\end{tabular}
\label{dataset_table}
\end{table}

% \begin{figure*}[h]
% \centering
% \includegraphics[width=0.9\linewidth]{result_zoomed_HAF.pdf}
% \caption{}
% \label{fig:result_zoomed_HAF}
% \end{figure*}

% \begin{figure*}[h]
% \centering
% \includegraphics[width=0.9\linewidth]{result_zoomed_VAF.pdf}
% \caption{Final output produced by LaneAF with vertical affinity fields; each color represents a unique lane instance based on affinity field decoding results.}
% \label{fig:result_zoomed_VAF}
% \end{figure*} 

\subsection{Metrics}

We use the same evaluation metrics used in past literature to make a representative comparison between our approach and prior work. This consists of the official metric of the TuSimple dataset (accuracy), the false positive (FP) rate, and the false negative (FN) rate. The TuSimple accuracy is calculated as:
\begin{equation}
Accuracy = \frac {N_{pred}}{N_{gt}}
\end{equation}
where ${N_{pred}}$ is the number of lane points that have been correctly predicted and ${N_{gt}}$ is the number of ground-truth lane points. 

Additionally, we report the $F1$ measure, which is based on the intersection over union (IoU) and is the only metric for CULane. This is calculated as in \cite{pan2018spatial}: 
\begin{equation}
F1 = 2 \cdot\left( \frac {precision \cdot recall}{precision + recall} \right)
\end{equation}
 where $precision$ is defined as $\frac{TP}{TP + FP}$, $recall$ is defined as $\frac{TP}{TP + FN}$, $TP$ is the number of lane points that have been correctly predicted, $FP$ is the number of false positives, and $FN$ is the number of false negatives.  This same $F1$ measure is also used for the lane approximations benchmark of the LLAMAS dataset. 

\subsection{Ablation Experiments}

In this subsection, we conduct a series of ablation experiments to validate our design choices. All ablation studies were conducted on the TuSimple validation set and can be seen in Table \ref{tab:ablation}. The first row contains the results of the standard LaneAF model, which we denote as the baseline model B. First, we train variants without the IoU loss (B w/o IoU) and the weighted binary cross-entropy loss (B w/o wBCE). Removing these losses decreased accuracy quite drastically while increasing the false positive and false negative rate. In fact, without the weighted binary cross-entropy loss, the F1 score in particular dropped significantly. The same is observed for the baseline model without random transformations during training (B w/o RT), as depicted in the fourth row. 

With regards to the down-sampling factor of the outputs, it is clear that the baseline model's factor of 4 achieved the best results; decreasing it to 2 (B (DS-2)) increased runtime and worsened accuracy and F1 slightly, while increasing it to 8 (B (DS-8)) had the most damaging effect on accuracy out of all modifications. We also trained a variant with 128 channels in the output head (B (HC-128)) compared to the original 256, and while this change had the smallest impact with respect to accuracy, it is evident that the baseline's 256 channels yields superior results. Finally, to validate the benefits of our clustering approach over standard multi-class segmentation, we trained a DLA-34 model to directly perform multi-class segmentation of all lanes (DLA-34 multi-class). This model obtained the worst F1 and accuracy scores out of all variants. This result clearly illustrates the effectiveness of binary segmentation followed by a separate affinity field-based clustering approach.

\begin{table}[t]
\centering
\resizebox{0.95\linewidth}{!}{%
\begin{threeparttable}\centering
\caption{LaneAF ablation experiments on the TuSimple validation set}
\label{tab:ablation}
\begin{tabular}{@{}ccccc@{}}
\hline
\begin{tabular}[c]{@{}c@{}}Model type\end{tabular} & 
\begin{tabular}[c]{@{}c@{}}F1 (\%)\end{tabular} & 
\begin{tabular}[c]{@{}c@{}}Acc (\%)\end{tabular} & 
\begin{tabular}[c]{@{}c@{}}FP\end{tabular} & 
\begin{tabular}[c]{@{}c@{}}FN\end{tabular}\\
\hline \hline
\rowcolor[HTML]{EFEFEF}
B\tnote{1}                   & 95.31 & 94.62 & 0.0435 & 0.0500 \\
B w/o IoU\tnote{2}           & 95.17 & 94.31 & 0.0456 & 0.0507 \\
\rowcolor[HTML]{EFEFEF}
B w/o wBCE\tnote{3}          & 93.75 & 94.19 & 0.0598 & 0.0649 \\
B w/o RT\tnote{4}            & 93.56 & 94.29 & 0.0614 & 0.0670 \\
\rowcolor[HTML]{EFEFEF}
B (DS-2)\tnote{5}            & 94.76 & 94.22 & 0.0484 & 0.0559 \\
B (DS-8)                     & 93.94 & 92.73 & 0.0549 & 0.0656 \\
\rowcolor[HTML]{EFEFEF}
B (HC-128)\tnote{6}          & 94.80 & 94.56 & 0.0503 & 0.0535 \\
DLA-34 multi-class\tnote{7}  & 88.86 & 92.59 & 0.1115 & 0.1114 \\
\bottomrule
\end{tabular}
\begin{tablenotes}
    \item[1] B: baseline model with down-sampling factor 4 and 256 channels in output head
    \item[2] IoU: IoU loss
    \item[3] wBCE: weighted BCE loss
    \item[4] RT: random transformations during training
    \item[5] DS-x: down-sampling factor for outputs
    \item[6] HC-x: number of channels in output head
    \item[7] DLA-34 multi-class: DLA-34 model trained for lane detection via multi-class segmentation
 \end{tablenotes}
\end{threeparttable}
}
\end{table}

\begin{table}[t]
\centering
\caption{LaneAF results on the TuSimple benchmark}
\label{accuracy_table}
\resizebox{0.99\linewidth}{!}{%
\begin{tabular}{@{}cccccc@{}}
	\hline
	Method & F1 ($\%$) & Acc ($\%$) & FP & FN & MACs (G)\\
	\hline\hline
	\rowcolor[HTML]{EFEFEF}
	ResNet-18 \cite{resnet} & 87.87 & 92.69 & 0.0948 & 0.0822 & -\\
	PolyLaneNet \cite{tabelini2021polylanenet} & 90.62 & 93.36 & 0.0942 & 0.0933 & \textbf{1.7}\\
	\rowcolor[HTML]{EFEFEF}
	Cascaded-CNN \cite{cascnn}& 90.82 & 95.24 & 0.1197 & 0.0620 & -\\
	LaneNet \cite{post} & 94.80 & 96.38 & 0.0780 & 0.0244 & -\\
	\rowcolor[HTML]{EFEFEF}
	ENet-SAD \cite{sad} & 95.92 & 96.64 & 0.0602 & 0.0205 & -\\
	SCNN \cite{pan2018spatial} & 96.53 & 96.53 & 0.0617 & \textbf{0.0180} & -\\
    % ENet \cite{sad} & 91.9611 & 93.02 0.0886 0.0734 & -\\
    \rowcolor[HTML]{EFEFEF}
	ResNet-34 \cite{resnet} & 96.77 & 92.84 & 0.0918 &0.0796 & -\\
	PINet \cite{ko2021key} & \textbf{97.20} & \textbf{96.75} & 0.0310 & 0.0250 & -\\
	\rowcolor[HTML]{EFEFEF}
	LaneATT(ResNet-18) \cite{tabelini2020keep} & 96.71 & 95.57 & 0.0356 & 0.0301 & 9.3\\
	LaneATT (ResNet-34) \cite{tabelini2020keep} & 96.77 & 95.63 & 0.0353 & 0.0292 & 18.0\\
	\rowcolor[HTML]{EFEFEF}
	LaneATT (ResNet-122) \cite{tabelini2020keep} & 96.06 & 96.10 & 0.0564 & 0.0217 & 70.5\\
	\textbf{LaneAF (DLA-34)} & 96.49 & 95.62 & \textbf{0.0280} & 0.0418 & 22.2\\ 
	\hline
\end{tabular}%
}
\end{table}

\subsection{Results}

Performance results from LaneAF on the TuSimple benchmark are shown in Table~\ref{accuracy_table}. It can be seen that our false positive rate sets a new standard (0.0280) among the current state-of-the-art. This demonstrates that our model does not incorrectly detect a lane pixel as often as other networks and that LaneAF's multi-branch approach leads to confident lane pixel predictions. While we obtain superior accuracy to other backbone architectures such as ResNet-18 and -34 \cite{resnet}, our approach falls slightly short of current state-of-the-art models such as PINet \cite{ko2021key}, ENet-SAD \cite{sad}, and SCNN \cite{pan2018spatial}. However, our false negative rate is only marginally higher, signifying that the incorrectly classified lane pixels are most likely at the very ends of the lanes. Additionally, six training runs were conducted on this dataset with different random seeds, producing a standard deviation of 0.12 on the accuracy metric. From the consistency of these results, we can see that our proposed method is robust.

Table~\ref{accuracy_table2} displays the state-of-the-art results of our model on the CULane benchmark. With this significantly larger and more complex dataset, we can see that LaneAF's performance improves greatly with respect to other models and demonstrates our network's ability to generalize. LaneAF (with DLA-34) outperforms the current state-of-the-art with an F1 score of 77.41\%, surpassing models of similar size and even LaneATT \cite{tabelini2020keep} with its largest backbone, ResNet-122. Moreover, LaneAF sets a new benchmark in a majority of categories, including difficult ones such as Dazzle, Shadow, No line, Curve, and Night, exhibiting our model's high adaptability to curving roads and challenging lighting conditions. 

\begin{table*}[t]
\centering
\caption{LaneAF state-of-the-art results on the CULane benchmark}
\label{accuracy_table2}
\resizebox{0.99\linewidth}{!}{%
\begin{tabular}{@{}cccccccccccc@{}}
	\hline
	Method & Total & Normal & Crowded & Dazzle & Shadow & No line & Arrow &  Curve & Cross & Night & MACs (G) \\
	\hline\hline
	\rowcolor[HTML]{EFEFEF}
 	ResNet-18 \cite{resnet} & 68.40 & 87.70 & 66.00 & 58.40 & 62.80 & 40.20 & 81.00 & 57.90 & 1743 & 62.10 & - \\
	ENet-SAD \cite{sad} & 70.80 & 90.10 & 68.80 & 60.20 & 65.90 & 41.60 & 84.00 & 65.70 & 1998 & 66.00 & - \\
 	\rowcolor[HTML]{EFEFEF}
	SCNN \cite{pan2018spatial} & 71.60 & 90.60 & 69.70 & 58.50 & 66.90 & 43.40 & 84.10 & 64.40 & 1990 & 66.10 & - \\
 	ResNet-34 \cite{resnet} & 72.30 & 90.70 & 70.20 & 59.50 & 69.30 & 44.40 & 85.70 & 69.50 & 2037 & 66.70 & - \\
	\rowcolor[HTML]{EFEFEF}
	ERFNet-Intra-KD \cite{hou2020inter} & 72.40 & - & - & - & - & - & - & - & - & - & - \\
	CurveLanes-NAS-M \cite{li2020curvelane} & 73.50 & 90.20 & 70.50 & 65.90 & 69.30 & 48.80 & 85.70 &  67.50 & 2359 & 68.20 & 33.7 \\
	\rowcolor[HTML]{EFEFEF}
	SIM-CycleGAN \cite{liu2020lane} & 73.90 & 91.80 & 71.80 & 66.40 & 76.20 & 46.10 & 87.80 & 67.10 & 2346 & 69.40 & - \\
	ERFNet-E2E \cite{yoo2020end} & 74.00 & 91.00 & 73.10 & 64.50 & 74.10 & 46.60 & 85.80 & 71.90 & 2022 & 67.90 & - \\
	\rowcolor[HTML]{EFEFEF}
	PINet \cite{ko2021key} & 74.40 & 90.30 & 72.30 & 66.30 & 68.40 & 49.80 & 83.70 & 65.60 & 1427 & 67.70 & - \\
    LaneATT (ResNet-18) \cite{tabelini2020keep} & 75.09 & 91.11 & 72.96 & 65.72 & 70.91 & 48.35 & 85.49 & 63.37 & \textbf{1170} & 68.95 & 9.3 \\
    \rowcolor[HTML]{EFEFEF}
    RESA-50 \cite{zheng2021resa} & 75.30 & 92.10 & 73.10 & 69.20 & 72.80 & 47.70 & 88.30 & 70.30 & 1503 & 69.9 & - \\
	LaneATT (ResNet-34) \cite{tabelini2020keep} & 76.68 & \textbf{92.14} & 75.03 & 66.47 & 78.15 & 49.39 & \textbf{88.38} & 67.72 & 1330 & 70.72 & 18.0 \\
	\rowcolor[HTML]{EFEFEF}
	LaneATT (ResNet-122) \cite{tabelini2020keep} & 77.02 & 91.74 & \textbf{76.16} & 69.47 & 76.31 & 50.46 & 86.29 & 64.05 & 1264 & 70.81 & 70.5 \\
% 	\bf{LaneAF (DLA-34)} & 77.25 & \textbf{92.18} & 75.42 & 71.56 & 78.82 & 51.05 & 87.29 & \textbf{71.91} & 1500 & 72.69 & 18.6 \\
	\bf{LaneAF (ENet)} & 74.24 & 90.12 & 72.19 & 68.70 & 76.34 & 49.13 & 85.13 & 64.40 & 1934 & 68.67 & \textbf{2.2} \\
	\rowcolor[HTML]{EFEFEF}
	\bf{LaneAF (ERFNet)} & 75.63 & 91.10 & 73.32 & 69.71 & 75.81 & 50.62 & 86.86 & 65.02 & 1844 & 70.90 & 22.2 \\
	\bf{LaneAF (DLA-34)} & \textbf{77.41} & 91.80 & 75.61 & \textbf{71.78} & \textbf{79.12} & \textbf{51.38} & 86.88 & \textbf{72.70} & 1360 & \textbf{73.03} & 23.6 \\
	\hline
\end{tabular}
}
\end{table*}

\begin{table}[t]
\centering
\caption{LaneAF results on the LLAMAS benchmark}
\label{accuracy_table3}
\resizebox{0.99\linewidth}{!}{%
\begin{tabular}{@{}cccccc@{}}
	\hline
	Method & F1 ($\%$) & Prec ($\%$) & Recall ($\%$)\\
	\hline\hline
	\rowcolor[HTML]{EFEFEF}
	PolyLaneNet \cite{tabelini2021polylanenet} & 88.40 & 88.87 & 87.93\\
	LaneATT(ResNet-18) \cite{tabelini2020keep} & 93.46 & \textbf{96.92} & 90.24\\
	\rowcolor[HTML]{EFEFEF}
	LaneATT (ResNet-34) \cite{tabelini2020keep} & 93.74 & 96.79 & 90.88\\
	LaneATT (ResNet-122) \cite{tabelini2020keep} & 93.54 & 96.82 & 90.47\\
	\rowcolor[HTML]{EFEFEF}
	\textbf{LaneAF (DLA-34)} & \textbf{96.07} & 96.91 & \textbf{95.26}\\ 
	\hline
\end{tabular}%
}
\end{table}

For the CULane dataset, we additionally trained LaneAF models with ENet \cite{enet} and ERFNet \cite{romera2017erfnet} backbones, where we forego the final few upsampling/transposed convolution layers to ensure a downsampling factor of 4 (same as the DLA-34 variant). This allows us to make direct comparisons between our approach and other approaches using the same backbone architecture. For instance, LaneAF with the ENet backbone outperforms ENet-SAD \cite{sad} by over 3\% with respect to the F1 score. When using ERFNet as the backbone network, LaneAF's F1 score eclipses other ERFNet-based models such as ERFNet-E2E \cite{yoo2020end} and ERFNet-Intra-KD \cite{hou2020inter} by 1.63\% and 3.23\%, respectively. These comparisons confirm that the performance gains of LaneAF are achieved through a combination of the DLA-34 backbone and our proposed affinity fields based clustering.

Furthermore, LaneAF again achieves state-of-the-art performance on the LLAMAS dataset with an F1 score of 96.07\%. This surpasses LaneATT's \cite{tabelini2020keep} best model by over 2\%, as shown in Table~\ref{accuracy_table3}. This gap in performance is due to LaneAF's high Recall score, which indicates that the model is more adept at retrieving true lane pixels. 

\begin{figure}[t]
\centering
\begin{subfigure}[b]{0.99\linewidth}
   \includegraphics[width=\linewidth]{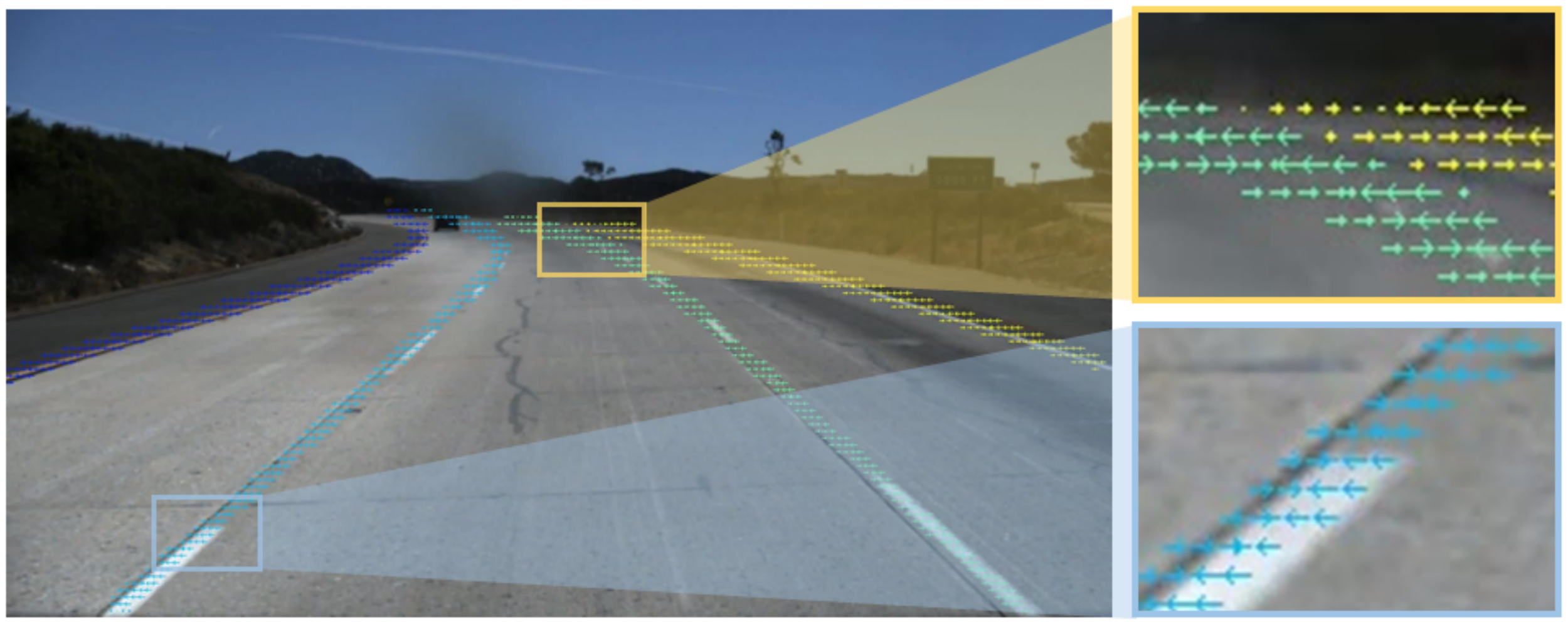}
   \caption{Predicted horizontal affinity field (HAF)}
   \label{fig:result_zoomed_HAF} 
\end{subfigure}

\begin{subfigure}[b]{0.99\linewidth}
   \includegraphics[width=\linewidth]{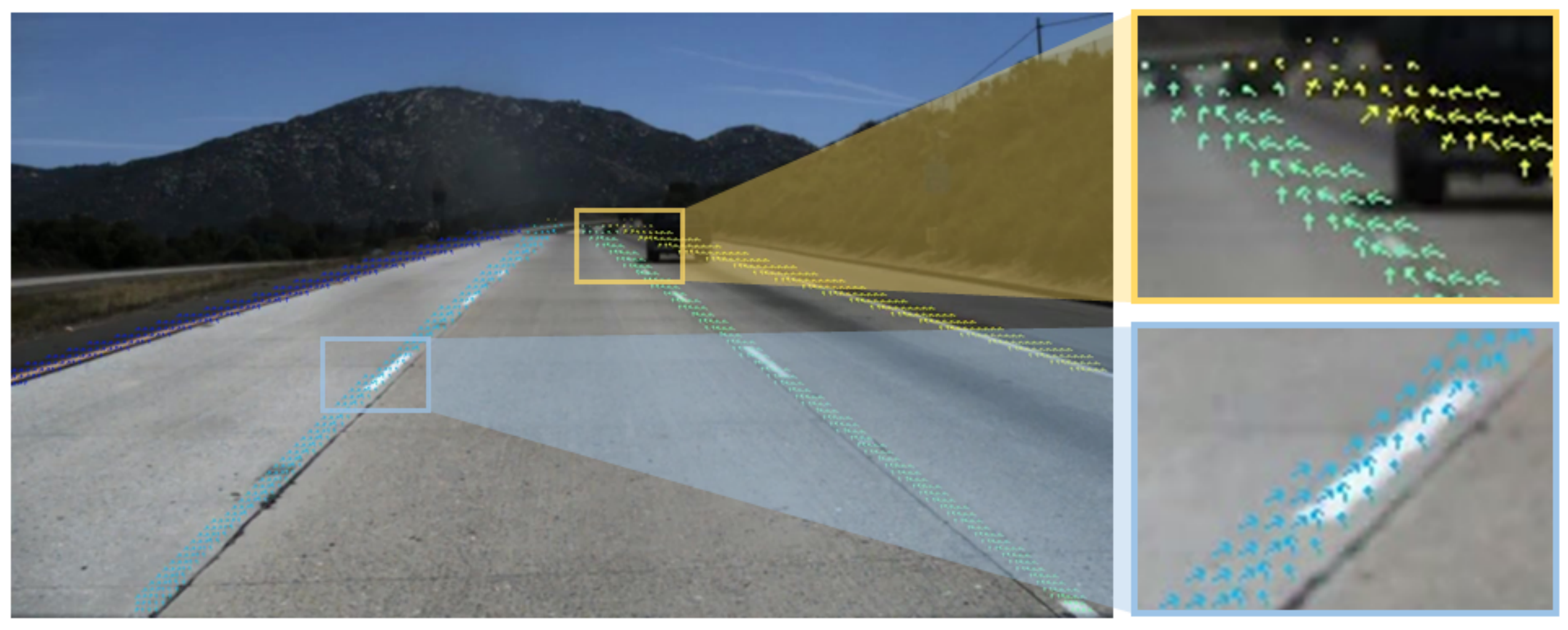}
   \caption{Predicted vertical affinity field (VAF)}
   \label{fig:result_zoomed_VAF}
\end{subfigure}
\caption{Example outputs produced by LaneAF with color coded affinity fields; each color represents a unique lane instance based on affinity field decoding. Of note is the successful discrimination of lane instances even as the lanes converge.}
\end{figure}

Qualitative examples of the predicted affinity fields from our approach are depicted in Figures~\ref{fig:result_zoomed_HAF} and \ref{fig:result_zoomed_VAF}. The clustered outputs shown here were created using the affinity field decoder, outlined in Algorithm \ref{alg:decoding}. In Figure \ref{fig:result_zoomed_HAF}, the HAF vectors point towards the center of their respective lane lines for each row of the output image. 
% This is based on the valid lane pixels found in the binary segmentation output and represents the locations of potential lane instances with respect to all detected lane pixels. 
Lane clusters are still successfully separated despite being closely located for numerous rows, demonstrated in yellow box of Figure \ref{fig:result_zoomed_HAF}. Likewise, in Figure \ref{fig:result_zoomed_VAF}, the VAF vectors point along the lane towards the mean location of the next row's lane pixels. This is visualized in the yellow box of Figure \ref{fig:result_zoomed_VAF}, where for each unique lane instance, the unit vector points towards the next row's mean lane pixel location. For both Figures \ref{fig:result_zoomed_HAF} and \ref{fig:result_zoomed_VAF}, the blue boxes clearly display how the HAF and VAF are implemented for a single detected lane instance. 

Another key point to note is the accuracy of the model at lane points that are farther away from the camera.
Since the ground truth segmentation masks for each lane are of approximately the same thickness in the image plane from top to bottom, the model is trained to predict thick foreground masks for lane points even if they are farther away. This results in little to no degradation for lane points that are far away. However, at the horizon, some clusters will be occasionally assigned to non-optimal lanes due to the close proximity of the lane lines.

\begin{figure}[t]
\centering
\includegraphics[width=0.99\linewidth]{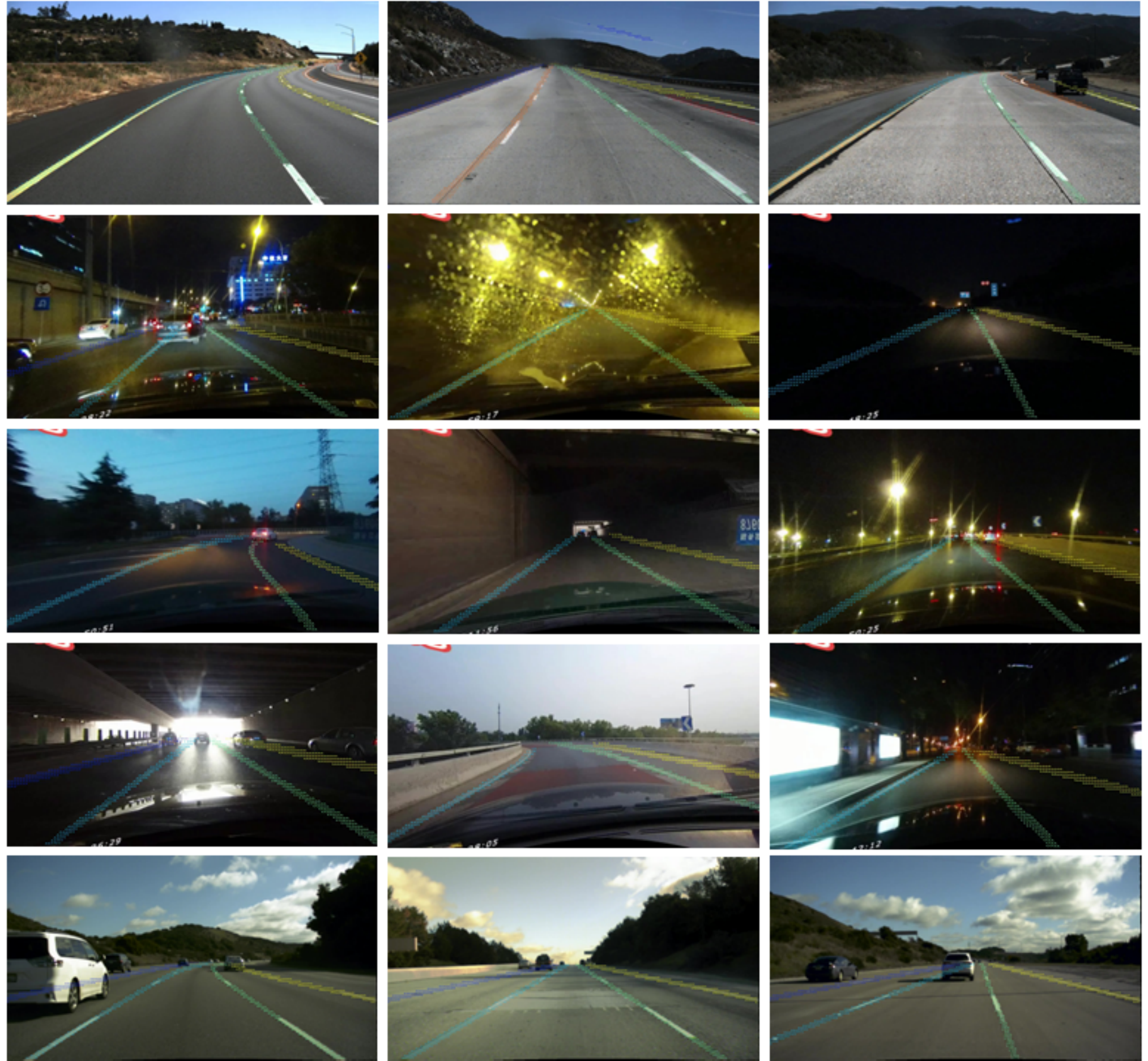}
\caption{LaneAF qualitative results on TuSimple (row 1), CULane (rows 2-4), and LLAMAS (row 5).}
\label{fig:result_grid}
\end{figure}

In Figure~\ref{fig:result_grid}, we show additional qualitative results from the TuSimple dataset (row 1), the CULane dataset (rows 2-4), and the LLAMAS dataset (row 4). The TuSimple examples demonstrate LaneAF's high performance on curved highways and on lanes that are merging and splitting due to entrances and exits, highlighting our model's flexibility to the number of lanes present on a given road. Also notable in the middle image of the first row is the false detection of a lane line due to an airplane contrail. The displayed results from CULane include challenging scenarios that illustrate LaneAF's robustness on curved roads and in very poor lighting conditions. This diverse set of examples exhibit characteristics of the Dazzle, Shadow, Curve, and Night categories of the CULane dataset. Finally, the LLAMAS samples show excellent performance on additional highway scenes similar to the TuSimple examples.

\section{Concluding Remarks}

In this paper, we proposed a novel approach to lane detection and instance segmentation through the use of binary segmentation masks and per-pixel affinity fields. The horizontal and vertical affinity fields, along with the predicted binary masks were demonstrated to successfully cluster lane pixels into unique lane instances in a post-processing step. This is accomplished using a simple row-by-row decoding process with little overhead, and enables LaneAF to detect a variable number of lanes of arbitrary width without assuming a fixed or maximum number of lanes. This form of clustering is also more interpretable in comparison to previous visual approaches since it can be analyzed to easily identify and correct sources of error. The ablation study conducted also validated the effectiveness of the approach over standard multi-class segmentation. Our proposed method achieves the lowest reported false positive rate (0.0280) on the TuSimple benchmark; on the larger and more comprehensive CULane dataset, LaneAF sets a new state-of-the-art result with a total F1 score of 77.41\%, surpassing much deeper and more complex models. LaneAF also achieves a state-of-the-art F1 score on the LLAMAS benchmark by a significant margin (+2\%), underscoring its robust performance. 

\section{Acknowledgements}
We are grateful to the reviewers for their valuable comments. We also thank our sponsors and colleagues at LISA, UC San Diego for their support.

%------------------------------------------------------------------------
%\section*{References}
{\small
\bibliographystyle{IEEEtran}
\bibliography{paper}

\begin{thebibliography}{10}
\providecommand{\url}[1]{#1}
\csname url@rmstyle\endcsname
\providecommand{\newblock}{\relax}
\providecommand{\bibinfo}[2]{#2}
\providecommand\BIBentrySTDinterwordspacing{\spaceskip=0pt\relax}
\providecommand\BIBentryALTinterwordstretchfactor{4}
\providecommand\BIBentryALTinterwordspacing{\spaceskip=\fontdimen2\font plus
\BIBentryALTinterwordstretchfactor\fontdimen3\font minus
  \fontdimen4\font\relax}
\providecommand\BIBforeignlanguage[2]{{%
\expandafter\ifx\csname l@#1\endcsname\relax
\typeout{** WARNING: IEEEtran.bst: No hyphenation pattern has been}%
\typeout{** loaded for the language `#1'. Using the pattern for}%
\typeout{** the default language instead.}%
\else
\language=\csname l@#1\endcsname
\fi
#2}}

\bibitem{daily2017self}
M.~Daily, S.~Medasani, R.~Behringer, and M.~Trivedi, ``Self-driving cars,''
  \emph{Computer}, vol.~50, no.~12, pp. 18--23, 2017.

\bibitem{deo2018would}
N.~Deo, A.~Rangesh, and M.~M. Trivedi, ``How would surround vehicles move? a
  unified framework for maneuver classification and motion prediction,''
  \emph{IEEE Transactions on Intelligent Vehicles}, vol.~3, no.~2, pp.
  129--140, 2018.

\bibitem{affinity}
Z.~Cao, T.~Simon, S.-E. Wei, and Y.~Sheikh, ``Realtime multi-person 2d pose
  estimation using part affinity fields,'' in \emph{Proceedings of the IEEE
  conference on computer vision and pattern recognition}, 2017, pp. 7291--7299.

\bibitem{yuen2019looking}
K.~Yuen and M.~M. Trivedi, ``Looking at hands in autonomous vehicles: A convnet
  approach using part affinity fields,'' \emph{IEEE Transactions on Intelligent
  Vehicles}, vol.~5, no.~3, pp. 361--371, 2019.

\bibitem{dla}
F.~Yu, D.~Wang, E.~Shelhamer, and T.~Darrell, ``Deep layer aggregation,'' in
  \emph{Proceedings of the IEEE conference on computer vision and pattern
  recognition}, 2018, pp. 2403--2412.

\bibitem{satzoda2015enhancing}
R.~K. Satzoda and M.~M. Trivedi, ``On enhancing lane estimation using
  contextual cues,'' \emph{IEEE Transactions on Circuits and Systems for Video
  Technology}, vol.~25, no.~11, pp. 1870--1881, 2015.

\bibitem{enet}
A.~Paszke, A.~Chaurasia, S.~Kim, and E.~Culurciello, ``Enet: A deep neural
  network architecture for real-time semantic segmentation,'' \emph{arXiv
  e-prints}, pp. arXiv--1606, 2016.

\bibitem{sensors}
R.~K. Satzoda and M.~M. Trivedi, ``Drive analysis using vehicle dynamics and
  vision-based lane semantics,'' \emph{IEEE Transactions on Intelligent
  Transportation Systems}, vol.~16, no.~1, pp. 9--18, 2014.

\bibitem{robust}
Q.~Zou, H.~Jiang, Q.~Dai, Y.~Yue, L.~Chen, and Q.~Wang, ``Robust lane detection
  from continuous driving scenes using deep neural networks,'' \emph{IEEE
  transactions on vehicular technology}, vol.~69, no.~1, pp. 41--54, 2019.

\bibitem{gan}
M.~Ghafoorian, C.~Nugteren, N.~Baka, O.~Booij, and M.~Hofmann, ``El-gan:
  Embedding loss driven generative adversarial networks for lane detection,''
  in \emph{Proceedings of the European Conference on Computer Vision (ECCV)
  Workshops}, 2018, pp. 0--0.

\bibitem{pan2018spatial}
X.~Pan, J.~Shi, P.~Luo, X.~Wang, and X.~Tang, ``Spatial as deep: Spatial cnn
  for traffic scene understanding,'' in \emph{Proceedings of the AAAI
  Conference on Artificial Intelligence}, vol.~32, no.~1, 2018.

\bibitem{fastDraw}
J.~Philion, ``Fastdraw: Addressing the long tail of lane detection by adapting
  a sequential prediction network,'' in \emph{Proceedings of the IEEE/CVF
  Conference on Computer Vision and Pattern Recognition}, 2019, pp.
  11\,582--11\,591.

\bibitem{leastSquare}
W.~Van~Gansbeke, B.~De~Brabandere, D.~Neven, M.~Proesmans, and L.~Van~Gool,
  ``End-to-end lane detection through differentiable least-squares fitting,''
  in \emph{Proceedings of the IEEE/CVF International Conference on Computer
  Vision Workshops}.\hskip 1em plus 0.5em minus 0.4em\relax IEEE, 2019, pp.
  905--913.

\bibitem{sad}
Y.~Hou, Z.~Ma, C.~Liu, and C.~C. Loy, ``Learning lightweight lane detection
  cnns by self attention distillation,'' in \emph{Proceedings of the IEEE/CVF
  International Conference on Computer Vision}, 2019, pp. 1013--1021.

\bibitem{post}
D.~Neven, B.~De~Brabandere, S.~Georgoulis, M.~Proesmans, and L.~Van~Gool,
  ``Towards end-to-end lane detection: an instance segmentation approach,'' in
  \emph{2018 IEEE intelligent vehicles symposium (IV)}.\hskip 1em plus 0.5em
  minus 0.4em\relax IEEE, 2018, pp. 286--291.

\bibitem{cascnn}
F.~Pizzati, M.~Allodi, A.~Barrera, and F.~Garc{\'\i}a, ``Lane detection and
  classification using cascaded cnns,'' in \emph{International Conference on
  Computer Aided Systems Theory}.\hskip 1em plus 0.5em minus 0.4em\relax
  Springer, 2019, pp. 95--103.

\bibitem{ko2021key}
Y.~Ko, Y.~Lee, S.~Azam, F.~Munir, M.~Jeon, and W.~Pedrycz, ``Key points
  estimation and point instance segmentation approach for lane detection,''
  \emph{IEEE Transactions on Intelligent Transportation Systems}, 2021.

\bibitem{bai2017deep}
M.~Bai and R.~Urtasun, ``Deep watershed transform for instance segmentation,''
  in \emph{Proceedings of the IEEE Conference on Computer Vision and Pattern
  Recognition}, 2017, pp. 5221--5229.

\bibitem{uhrig2016pixel}
J.~Uhrig, M.~Cordts, U.~Franke, and T.~Brox, ``Pixel-level encoding and depth
  layering for instance-level semantic labeling,'' in \emph{German Conference
  on Pattern Recognition}.\hskip 1em plus 0.5em minus 0.4em\relax Springer,
  2016, pp. 14--25.

\bibitem{mulFeature}
T.~Gupta, H.~S. Sikchi, and D.~Charkravarty, ``Robust lane detection using
  multiple features,'' in \emph{2018 IEEE Intelligent Vehicles Symposium
  (IV)}.\hskip 1em plus 0.5em minus 0.4em\relax IEEE, 2018, pp. 1470--1475.

\bibitem{end}
N.~Garnett, R.~Cohen, T.~Pe'er, R.~Lahav, and D.~Levi, ``3d-lanenet: end-to-end
  3d multiple lane detection,'' in \emph{Proceedings of the IEEE/CVF
  International Conference on Computer Vision}, 2019, pp. 2921--2930.

\bibitem{tabelini2020keep}
L.~Tabelini, R.~Berriel, T.~M. Paixao, C.~Badue, A.~F. De~Souza, and
  T.~Oliveira-Santos, ``Keep your eyes on the lane: Real-time attention-guided
  lane detection,'' in \emph{Proceedings of the IEEE/CVF Conference on Computer
  Vision and Pattern Recognition}, 2021, pp. 294--302.

\bibitem{spatiTemporal}
Y.~Huang, S.~Chen, Y.~Chen, Z.~Jian, and N.~Zheng, ``Spatial-temproal based
  lane detection using deep learning,'' in \emph{IFIP International conference
  on artificial Intelligence applications and innovations}.\hskip 1em plus
  0.5em minus 0.4em\relax Springer, 2018, pp. 143--154.

\bibitem{snesor}
M.~Bai, G.~Mattyus, N.~Homayounfar, S.~Wang, S.~K. Lakshmikanth, and
  R.~Urtasun, ``Deep multi-sensor lane detection,'' in \emph{2018 IEEE/RSJ
  International Conference on Intelligent Robots and Systems (IROS)}.\hskip 1em
  plus 0.5em minus 0.4em\relax IEEE, 2018, pp. 3102--3109.

\bibitem{resnet}
K.~He, X.~Zhang, S.~Ren, and J.~Sun, ``Deep residual learning for image
  recognition,'' in \emph{Proceedings of the IEEE conference on computer vision
  and pattern recognition}, 2016, pp. 770--778.

\bibitem{romera2017erfnet}
E.~Romera, J.~M. Alvarez, L.~M. Bergasa, and R.~Arroyo, ``Erfnet: Efficient
  residual factorized convnet for real-time semantic segmentation,'' \emph{IEEE
  Transactions on Intelligent Transportation Systems}, vol.~19, no.~1, pp.
  263--272, 2017.

\bibitem{huang2017densely}
G.~Huang, Z.~Liu, L.~Van Der~Maaten, and K.~Q. Weinberger, ``Densely connected
  convolutional networks,'' in \emph{Proceedings of the IEEE conference on
  computer vision and pattern recognition}, 2017, pp. 4700--4708.

\bibitem{dai2017deformable}
J.~Dai, H.~Qi, Y.~Xiong, Y.~Li, G.~Zhang, H.~Hu, and Y.~Wei, ``Deformable
  convolutional networks,'' in \emph{Proceedings of the IEEE international
  conference on computer vision}, 2017, pp. 764--773.

\bibitem{llamas}
K.~Behrendt and R.~Soussan, ``Unsupervised labeled lane markers using maps,''
  in \emph{Proceedings of the IEEE/CVF International Conference on Computer
  Vision Workshops}, 2019, pp. 0--0.

\bibitem{tabelini2021polylanenet}
L.~Tabelini, R.~Berriel, T.~M. Paixao, C.~Badue, A.~F. De~Souza, and
  T.~Oliveira-Santos, ``Polylanenet: Lane estimation via deep polynomial
  regression,'' in \emph{2020 25th International Conference on Pattern
  Recognition (ICPR)}.\hskip 1em plus 0.5em minus 0.4em\relax IEEE, 2021, pp.
  6150--6156.

\bibitem{hou2020inter}
Y.~Hou, Z.~Ma, C.~Liu, T.-W. Hui, and C.~C. Loy, ``Inter-region affinity
  distillation for road marking segmentation,'' in \emph{Proceedings of the
  IEEE/CVF Conference on Computer Vision and Pattern Recognition}, 2020, pp.
  12\,486--12\,495.

\bibitem{li2020curvelane}
Z.~Li, ``Curvelane-nas: Unifying lane-sensitive architecture search and
  adaptive point blending,'' in \emph{Proceedings of the European Conference on
  Computer Vision (ECCV)}.\hskip 1em plus 0.5em minus 0.4em\relax Springer,
  2020.

\bibitem{liu2020lane}
T.~Liu, Z.~Chen, Y.~Yang, Z.~Wu, and H.~Li, ``Lane detection in low-light
  conditions using an efficient data enhancement: Light conditions style
  transfer,'' in \emph{2020 IEEE Intelligent Vehicles Symposium (IV)}.\hskip
  1em plus 0.5em minus 0.4em\relax IEEE, 2020, pp. 1394--1399.

\bibitem{yoo2020end}
S.~Yoo, H.~S. Lee, H.~Myeong, S.~Yun, H.~Park, J.~Cho, and D.~H. Kim,
  ``End-to-end lane marker detection via row-wise classification,'' in
  \emph{Proceedings of the IEEE/CVF Conference on Computer Vision and Pattern
  Recognition Workshops}, 2020, pp. 1006--1007.

\bibitem{zheng2021resa}
T.~Zheng, H.~Fang, Y.~Zhang, W.~Tang, Z.~Yang, H.~Liu, and D.~Cai, ``Resa:
  Recurrent feature-shift aggregator for lane detection,'' in \emph{Proceedings
  of the AAAI Conference on Artificial Intelligence}, vol.~35, no.~4, 2021, pp.
  3547--3554.

\end{thebibliography}
}

\end{document}